\documentclass[a4paper, twocolumn, 11pt]{article}
\usepackage[utf8]{inputenc} 
\usepackage{amssymb}
\usepackage{amsmath}
\usepackage{graphicx}
\usepackage{dcolumn}
\usepackage{hyperref}
\usepackage{subcaption}
\usepackage{tabularx} 
\usepackage{float}
\usepackage{stfloats}

\DeclareMathOperator{\sign}{sign}
\newcolumntype{Y}{>{\raggedleft\arraybackslash}X}

\begin{document}

\title{Ubicomp Digital 2020 - Handwriting classification using a convolutional recurrent network}
\author{Wei-Cheng Lai, Hendrik Schröter}

\maketitle

\begin{abstract}
The Ubicomp Digital 2020 - Time Series Classification Challenge from STABILO is a challenge about multi-variate time series classification. The data collected from 100 volunteer writers, and contains 15 features measured with multiple sensors on a pen. In this paper, we use a neural network to classify the data into 52 classes, that is lower and upper cases of Arabic letters. The proposed architecture of the neural network a is CNN-LSTM network. It combines convolutional neural network (CNN) for short term context with a long short-term memory layer (LSTM) for also long-term dependencies. We reached an accuracy of 68\% on our writer exclusive test set and 64.6\% on the blind challenge test set resulting in the second place.
\end{abstract}

\section{Introduction}
Handwriting is an important skill, and it's being used actually daily in our life \cite{handwriting}. The research indicated that contemporary handwriting has influence on the developing capabilities of children, atypical development  \cite{primaryandintermediategrade, seventonineolds}. With the benefit of the digitization, handwriting is no more just off-line. Nowadays, there are many products, like tablets, which people can write on. They help us improve the life quality and make it easier to do meetings or tutorials online. Handwriting is a fine motor skills, for which various kinds of movement recording devices were employed \cite{digitialrecording}. Many methods for handwriting classification focuses on character recognition on pixel level. Therefore, the data was usually based on images \cite{characterrecognition}. In this paper, we recognize letters not based on images but based on sensor data recorded in a pen. This allows to write on a paper, while transmitting the digitized writings to a connected device such as a smartphone or computer. This technique can also eventually be used for autocorrection. Unlike traditional image data for handwriting recognition \cite{databasehandwrite}, the data for the challenge were collected from sensors, including two accelerometers, a gyroscope, a magnetometer and a force sensor. Therefore, we couldn't not use the way of image classification, but the way to solve the multivariate time series classification.

\section{Data}
\begin{figure}
\graphicspath{{images/}}
\includegraphics[width=0.5\textwidth]{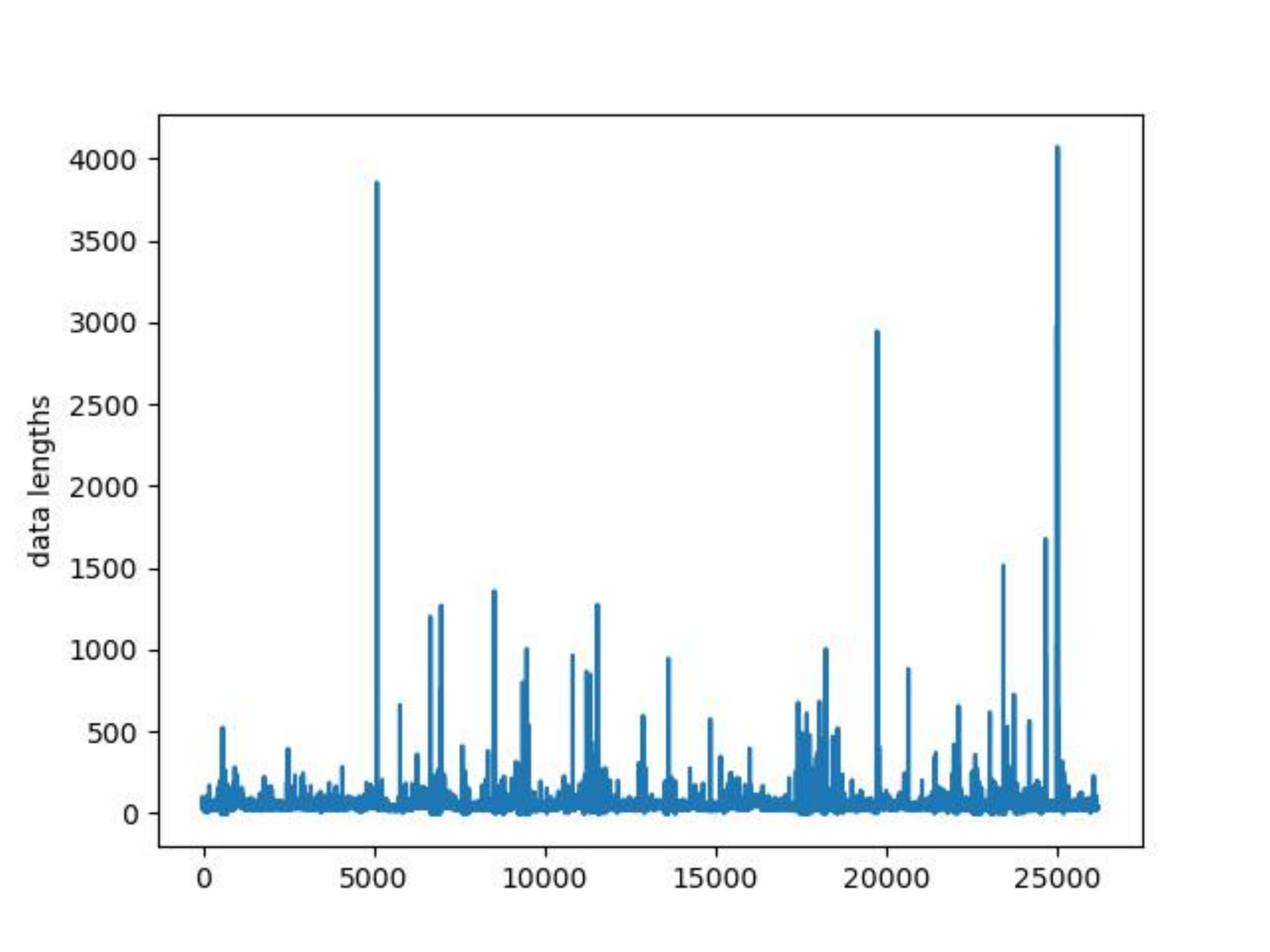}
\caption{\label{fig:histogram}Histogram of raw data length}
\end{figure}

The dataset provided by STABILO for the Ubicomp Digital 2020 challenge contains data from 100 writers. It was collected using a so called Digipen, a pen equipped with multiple sensors such as two 3-axis accelerometers (front and rear), a 3-axis gyroscope, a 3-axis magnetometer and force sensor sampled at 100~Hz. Each data sample needs a corresponding calibration to correct the bias and scale of each feature. The data includes upper case letters and lower case letters of the Arabic letters. Therefore, 52 classes are used for the classification task. All 100 volunteers were right-handed. There was no guideline for the way of holding the pen. The subjects held the pen in their natural way. The data lengths from the raw data can be viewed at Figure \ref{fig:histogram}.

\section{Methods}
\subsection{Data processing}
We separated the whole dataset with 80\% for training set, and 20\% for test set, which were separately 20979 samples and 5200 samples. The writers didn't overlap in training set and test set. This ensured that we preserve the writers' uniqueness in our dataset. Each sample had a different length. Before we entered the data into model, we needed to process them to the same length because of the input length of the convolutional neural network. We used the Fourier method to transcribe each signal to the same length. We also got rid of the features of 3-axis magnetometer, because we found that they don't have great influence on classification. The unit of 3-axis gyroscope features would also be transformed from angle to radian, in order to let the range of values become more similar to other features, At the end, we didn't use the traditional zero-mean normalization, but applied a logarithm scaling. We found that logarithmic transformation of the feature space, which results in smaller value ranges being resolved more finely than large ones.\newline

The logarithm formula:
\begin{equation}
    S_{\log}=\sign(S)\cdot\log(|S| + 1)\text{\ ,}
\end{equation}
where $S$ is the original feature vector.

\subsection{Model of deep learning network}
Based on the good performance of convolutional neural network on sentence-level classification tasks \cite{sentenceclassification}, we found that it would be helpful in letter classification task. Because of the information store in cells of long short-term memory (LSTM) \cite{lstm, Lai2015RecurrentCN}, we could also benefit from using it on time series data. Our final Model is based on a combination of convolutional neural network (CNN) and recurrent neural network (RNN), which was e.g. also used for language modeling \cite{cnnlstm}. There are 8 convolutional layers and a LSTM layer after the convolutional part in our proposed model. After the third convolutional layer, we did an exponential transformation, which transforms back to the normal value ranges. Between the convolution layers, we also applied the activation function (Rectified Linear Units), Batch normalization \cite{batchnormalization} and Max pooling layer. At the end, we took the values of hidden state of LSTM layer, which includes only the value of the last time state, and placed them into the fully connected layer, in order to predict the probabilities of 52 classes. In our proposed network, we used relative small learning rate $3\cdot10^{-5}$ and also weight decay $1\cdot10^{-3}$ for Adam optimizer \cite{adam}. Architecture of the model can be viewed at Table \ref{tab:architecture}.

\begin{table}
\centering
\begin{tabularx}{\linewidth}{l|Y}
Layer Name & CNN-LSTM-Net12 \\\hline
conv1 & 1 x 11, 32, stride 2  \\
conv2 & 1 x 11, 32, stride 1  \\
conv3 & 1 x 11, 32, stride 1  \\
exp   & exponential           \\
conv4 & 1 x 5, 64, stride 1  \\
maxpool1 & 1 x 7, stride 3  \\
conv5 & 1 x 5, 64, stride 1  \\
maxpool2 & 1 x 7, stride 3  \\
conv6 & 1 x 3, 128, stride 1  \\
conv8 & 1 x 3, 128, stride 1  \\
conv9 & 1 x 3, 128, stride 1  \\
lstm1 & hidden size 256, layer 1\\
fully connected & 256 x 52 fully connection\\
\end{tabularx}
\caption{Model Architecture}
\label{tab:architecture}
\end{table}

\begin{figure*}[hbt]
 \graphicspath{{images/}}
 \begin{minipage}[b]{.49\linewidth}
    \centering
    \includegraphics[width=1\linewidth]{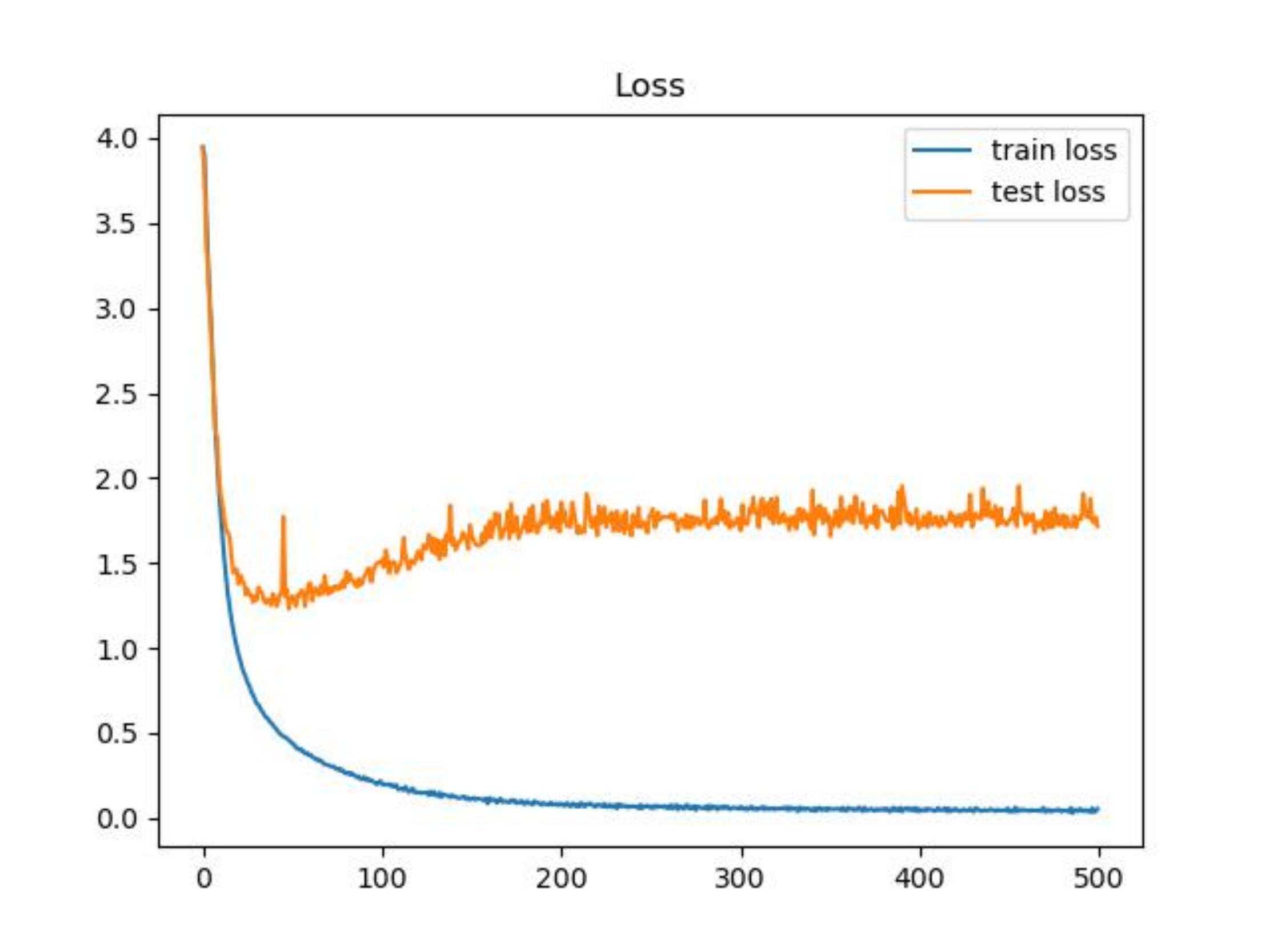}
    \subcaption{}\label{fig:Ng1}
 \end{minipage}
 \hfil
 \begin{minipage}[b]{.49\linewidth}
    \centering
    \includegraphics[width=1\linewidth]{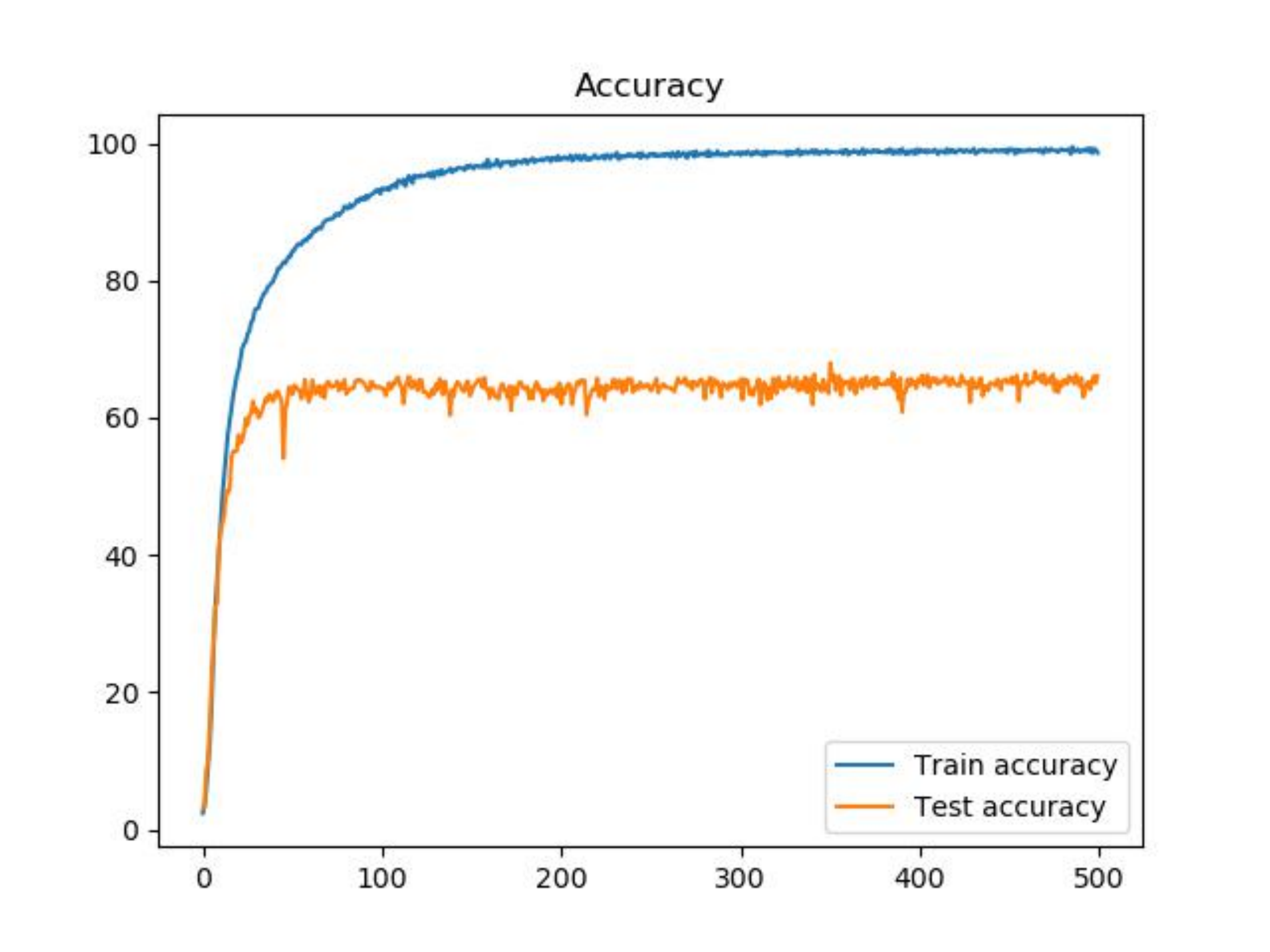}
    \subcaption{}\label{fig:Ng2}
 \end{minipage}
 \caption{Overview of the result from the proposed model. (a) Cross Entropy Loss, overfitting after 100 epochs. (b) Accuracy, reaching 68\% on the test set.}
 \label{fig:results}
\end{figure*}
\section{Results}
\begin{figure*}[hbt]
    \graphicspath{{images/}}
    \centering
    \includegraphics[width=1.1\linewidth]{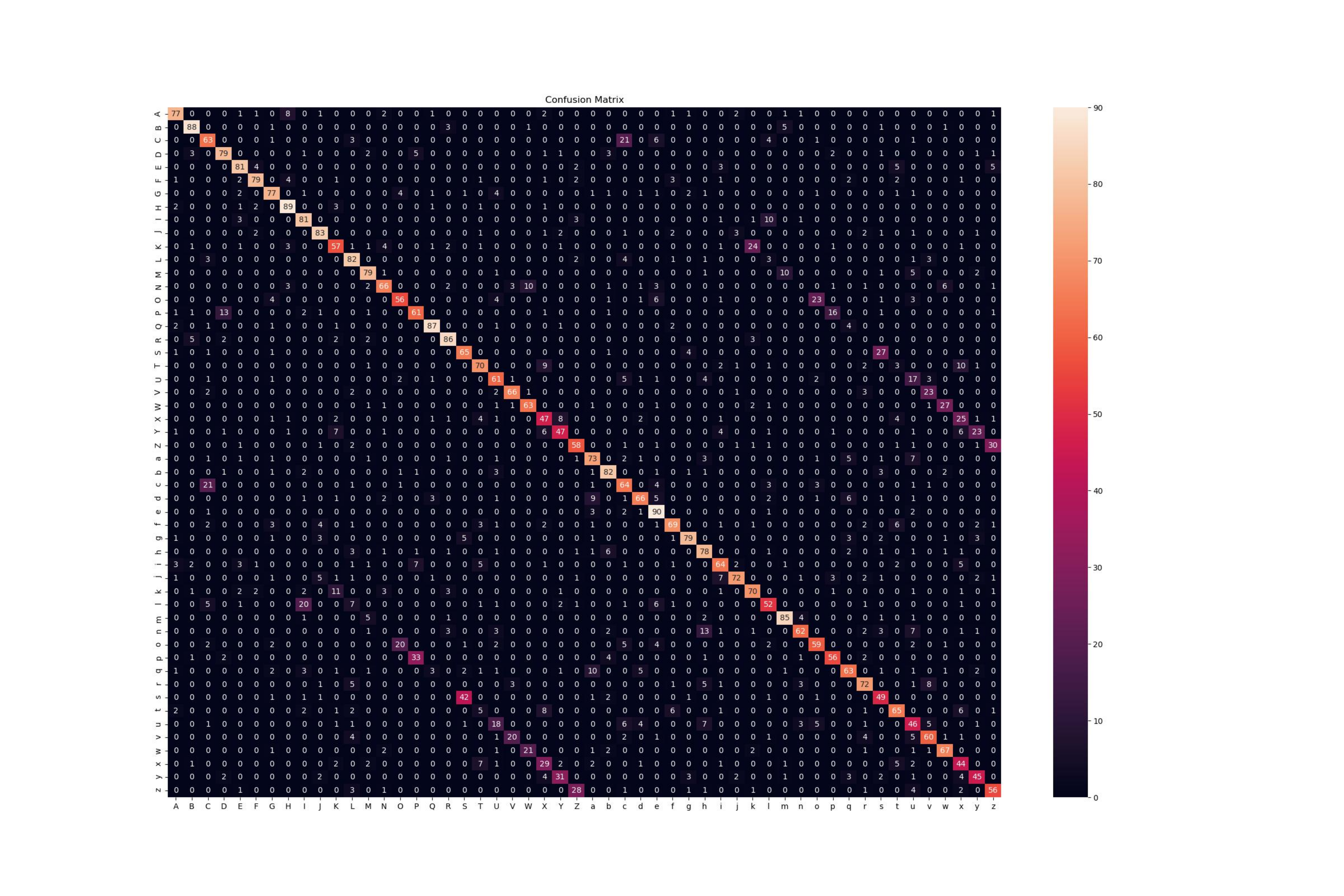}
    \caption{Confusion Matrix for the test set. Here, you can see the ambiguity sometimes between lower and upper case letters.}
    \label{fig:Ng3}
\end{figure*}

We have trained on our CNN-LSTM-Net12 model for 500 epochs resulting in an accuracy of 68\% on the test set. The Loss and Accuracy can be viewed at Figure \ref{fig:Ng1} and \ref{fig:Ng2}. Confusion Matrix can be viewed at Figure \ref{fig:Ng3}. We have also observed that the network has some difficulty to distinguish the lower and upper case of the same letters. This observation can be seen in Confusion Matrix. This problem especially holds for letters that are similar in the lower and upper cases. We believe that for a given word and sentence context, this is likely to improve, since upper case letters usually occur at the beginning of a sentence. In Figure \ref{fig:results}, we can also see that the network is actually overfitting. The test accuracy though does not decrease significantly while proceeding with the training. Thus, for submission, we retrained the model using the whole dataset without any evaluation on a test set. The resulting train accuracy of 99.3\% indicates that our proposed model could still model the whole training data distribution. Due to the overfitting characteristic our final submission is from epoch 150.

\section{Conclusion}
Our proposed model could classify correctly most of the letters from the dataset. Nevertheless, if the upper and lower case letters are too similar, our model wouldn't be able to recognize the correct letter. For example, letters like c, C or o, O, it's sometimes even not easy for human beings to distinguish the difference, as human beings look normally the whole sentence but not just a single letter. This problem can actually be solved in the future, if we have more data and may eventually have the sentence-based dataset for stage 3. There were actually two improvements on our model during the training process. First, we changed the unit of gyroscope features from angles to radian. Seconds, we added a recurrent neural network (LSTM) after the convolutional neural network. Each of them made the improvement 3-6\% for accuracy on test set. It was also interesting that Resnet18 \cite{resnet} didn't work as well as the sequential convolutional neural network. It was overfitting and couldn't fit well into test set.


\bibliographystyle{unsrt}
\bibliography{refs}

\end{document}